\documentclass[conference]{IEEEtran}
\IEEEoverridecommandlockouts
\usepackage{cite}
\usepackage{amsmath,amssymb,amsfonts}
\usepackage{algorithmic}
\usepackage{graphicx}
\usepackage{textcomp}
\usepackage{xcolor}
\usepackage{stmaryrd}
\usepackage{amsthm}
\usepackage{textcomp}
\usepackage{array,float,url,graphicx,wrapfig,verbatim,color,comment}
\usepackage[caption=false]{subfig}
\usepackage{bm}
\usepackage{algorithm}
\usepackage{amssymb}
\usepackage{pgfgantt}
\usepackage{multirow}
\usepackage{soul}
\usepackage{bbding}
\usepackage{siunitx}

\usepackage{enumerate}
\usepackage{booktabs}
\usepackage{tabularx}
\usepackage{bbm}
\usepackage{xr}

\usepackage{mathtools}
\usepackage{cases}
\MHInternalSyntaxOn
\renewcommand{\dcases}
 {
  \MT_start_cases:nnnn
    {\quad}
    {$\m@th\displaystyle##$\hfil}
    {$\m@th\displaystyle##$\hfil}
    {\lbrace}
 }
\MHInternalSyntaxOff
\newcounter{relctr} 
\everydisplay\expandafter{\the\everydisplay\setcounter{relctr}{0}} 

\AtBeginDocument{} 

\DeclareMathOperator*{\argmin}{arg\,min}

\def\BibTeX{{\rm B\kern-.05em{\sc i\kern-.025em b}\kern-.08em
    T\kern-.1667em\lower.7ex\hbox{E}\kern-.125emX}}
\begin{document}

\title{Heterogeneity-Aware Resource Allocation and Topology Design for Hierarchical Federated Edge Learning}

\author{Zhidong~Gao,~\IEEEmembership{Student~Member,~IEEE,} Yu~Zhang,~\IEEEmembership{Student~Member,~IEEE,} Yanmin~Gong,~\IEEEmembership{Senior~Member,~IEEE,}\\
and Yuanxiong~Guo,~\IEEEmembership{Senior~Member,~IEEE}
\IEEEcompsocitemizethanks{\IEEEcompsocthanksitem Z. Gao, Y. Zhang, and Y. Gong are with the Department of Electrical and Computer Engineering, The University of Texas at San Antonio, San Antonio, TX, 78249. E-mail: \{zhidong.gao@my., yu.zhang@my., yanmin.gong@\}utsa.edu

Y. Guo is with the Department of Information Systems and Cyber Security, The University of Texas at San Antonio, San Antonio, TX, 78249. E-mail: yuanxiong.guo@utsa.edu}
}



\maketitle

\begin{abstract}
Federated Learning (FL) provides a privacy-preserving framework for training machine learning models on mobile edge devices. Traditional FL algorithms, e.g., FedAvg, impose a heavy communication workload on these devices. To mitigate this issue, Hierarchical Federated Edge Learning (HFEL) has been proposed, leveraging edge servers as intermediaries for model aggregation. Despite its effectiveness, HFEL encounters challenges such as a slow convergence rate and high resource consumption, particularly in the presence of system and data heterogeneity. However, existing works are mainly focused on improving training efficiency for traditional FL, leaving the efficiency of HFEL largely unexplored. In this paper, we consider a two-tier HFEL system, where edge devices are connected to edge servers and edge servers are interconnected through peer-to-peer (P2P) edge backhauls. Our goal is to enhance the training efficiency of the HFEL system through strategic resource allocation and topology design. Specifically, we formulate an optimization problem to minimize the total training latency by allocating the computation and communication resources, as well as adjusting the P2P connections. To ensure convergence under dynamic topologies, we analyze the convergence error bound and introduce a model consensus constraint into the optimization problem. The proposed problem is then decomposed into several subproblems, enabling us to alternatively solve it online. Our method facilitates the efficient implementation of large-scale FL at edge networks under data and system heterogeneity. Comprehensive experiment evaluation on benchmark datasets validates the effectiveness of the proposed method, demonstrating significant reductions in training latency while maintaining the model accuracy compared to various baselines.

\end{abstract}









\section{Introduction}


\IEEEPARstart{T}{he} widespread adoption of edge devices, such as smartphones and Internet-of-things (IoT) devices, each possessing advanced sensing, computing, and storage capabilities, results in an enormous amount of data being produced daily at the network edge. Concurrently, the rapid advancements in artificial intelligence (AI) and machine learning (ML) facilitate the extraction of valuable knowledge from extensive data. The integration of 5G networks with AI/ML technologies is driving the development of numerous innovative applications that have significant economic and social impacts, such as autonomous driving~\cite{wu2023crossfuser}, augmented reality~\cite{xiong2021augmented}, real-time video analytics~\cite{li2020reducto}, mobile healthcare~\cite{wang2024analyzing}, and smart manufacturing~\cite{cai2024fedhip}. A key characteristic of these new applications is the substantial and continuously streaming data they produce, which requires efficient processing for real-time learning and decision-making. However, despite these advancements, data sharing is impeded by privacy regulations, such as the General Data Protection Regulation (GDPR), and hardware constraints, like limited communication bandwidth. Federated learning (FL)~\cite{mcmahan2017communication} emerges as a promising solution to these challenges by enabling model training directly on mobile devices in a decentralized manner. FL not only enhances privacy protection but also leverages the computational resources of edge devices.

In a standard FL architecture, the system comprises a cloud-based Parameter Server (PS) and multiple clients\footnote{Note that we use clients and mobile devices interchangeably in this paper.}. The PS orchestrates the training procedure while the clients carry out model training. The training process consists of multiple communication rounds between the clients and the PS. In each round, each selected client downloads the most recent global model from PS and updates the model using its local data. Then, these updated local models are uploaded to the PS, where they are aggregated into a new global model. However, the process of transferring models between the numerous clients and the PS imposes a considerable data traffic load, leading to increased training latency and network congestion.



To unlock the full potential of FL over mobile edge networks, recent works~\cite{sun2022semi,zhang2022scalable,saha2020fogfl,liu2020client,wang2022demystifying,castiglia2020multi} have explored Hierarchical Federated Edge Learning (HFEL) by leveraging multi-server collaboration for model training. These works typically adopt a hierarchical architecture that integrates the advantage of the cloud-based and decentralized federated learning (DFL)~\cite{lian2017can,hua2022efficient,yu2019parallel,li2019communication} to enhance the speed and reliability of the FL system. In these works, clients are connected to a proximal edge server via wireless networks, while edge servers either connect to a central cloud server via an edge-to-cloud (E2C) network or form a peer-to-peer (P2P) network without connecting to a central cloud server. Each edge server acts as a local PS, orchestrating the training process with its connected clients. To facilitate collaborative training over broader distances and ensure a unified model across the network, edge servers engage in periodic synchronization, sharing, and updating their models with each other via E2C or P2P networks.


However, implementing FL efficiently in realistic edge systems presents significant challenges due to two key factors. 1) System heterogeneity: Edge devices possess limited resources such as battery life, communication bandwidth, and CPU frequency. Additionally, these resources are not uniformly distributed among different devices. In traditional HFEL, the faster devices must remain idle, waiting for the slower ones to finish training in each communication round. 2) Statistical heterogeneity: The data that local devices collect are inherently influenced by their specific geographic locations or operational contexts, leading to a non-IID data distribution. This discrepancy can significantly hinder the model’s convergence speed and accuracy.


To deal with the challenges introduced by system and data heterogeneity in FL environments, recent research~\cite{jiang2022adaptive,wang2022federated,li2023anycostfl,wu2023joint} has studied optimization-based strategies to enhance the FL system efficiency. Typically, these approaches first construct an optimization problem that encapsulates resource cost, system constraints, and training performance. Then, an adaptive control strategy is developed by solving the optimization problem to optimize device resource allocation and learning task scheduling in real time. Specifically, to guarantee the performance of the learned model under devised control decisions, these studies usually conduct the convergence analysis for their learning algorithm. This allows them to derive several key insights, which are subsequently applied to the design or solution of their optimization problems. However, these studies primarily focus on cloud-based FL or DFL, and less effect has been made for the FL system with hierarchical architectures.


In this paper, we investigate the efficiency of a two-tier HFEL system, where edge devices are connected with proximate edge servers and edge servers are interconnected via P2P networks. Each edge server with its connected edge devices forms one cluster. Our exploration is centered on the strategic regulation of edge device CPU processing, wireless communication bandwidth, and the P2P network configuration to unveil potential efficiency gains. In particular, we formulate an optimization problem that characterizes the inherent resource constraints of the HFEL system and the relationship between resource consumption and training latency. Moreover, we introduce a consensus distance constraint within the proposed optimization problem. This constraint captures the interplay of topology design and model convergence, particularly in the context of heterogeneous system resources and data distributions. To solve the proposed problem, we decompose it into several subproblems. Then, we propose an online alternative algorithm, dubbed FedRT, to find the solution iteratively. These control decisions are dynamically adjusted based on real-time assessments of system state, environmental conditions, and available resources, ensuring an efficient HFEL system.


To validate the effectiveness of our proposed method, we conducted comprehensive evaluations on three benchmark datasets, CIFAR-10~\cite{krizhevsky2009learning}, FEMNIST~\cite{caldas2019leaf}, and FMNIST~\cite{caldas2019leaf}under various data distributions and resource configurations. The experimental results illustrate that our method outperforms baselines in terms of the total training latency and convergence speed. In summary, our main contributions are stated as follows:


\begin{itemize}
    \item We formulated an optimization problem that captures the two trade-offs for the HFEL system: i) model convergence rate versus network topology link density by limiting consensus distance during training. ii) total training latency versus energy cost by finding the optimal communication and computation resource allocation scenario for edge devices. 
    
    \item We propose FedRT, which jointly optimizes the edge backhaul topology and resource allocation for a two-tier HFEL system, considering both system and data heterogeneity. FedRT achieves high accuracy, low latency, and resource-efficient model training in mobile edge FL.

    \item We present a detailed analysis of our control strategy and provide several insights on adjusting the control variables to achieve a better trade-off between resource consumption and convergence speed.

    \item We conduct extensive experiments on benchmark datasets. The results demonstrate the advantage of our method compared with baselines without client scheduling and/or resource allocation. 
\end{itemize}

The remainder of this paper is organized as follows. 
Section~\ref{sec:sys_model} describes the system model and formulated optimization problem. Sectio~\ref{sec:related_work} discusses the related works. Section~\ref{sec:converge} presents the convergence result and discusses the insights of topology design. Section~\ref{sec:problem_solve} introduces the proposed solution. Then, in Section \ref{sec:exp}, we demonstrate the experiment results. 
Finally, we conclude this paper in Section~\ref{sec:conclusion}.

\section{Related Work}\label{sec:related_work}
FL at mobile edge networks faces several challenges, such as high training
latency and resource cost due to the presence of system and data heterogeneity. To mitigate these issues, several resource-optimization-based algorithms have been proposed to enhance both training latency and resource utilization of FL. For instance, Luo et al.~\cite{luo2022tackling} introduced an adaptive client sampling algorithm aimed at minimizing convergence time in heterogeneous systems. Perazzone et al.~\cite{Peraz2022commun} proposed a joint client sampling and power allocation scheme to reduce convergence error and average communication time under a power constraint, though their approach does not consider local computation. Wang et al.~\cite{wang2023federated} formulated an optimization problem to minimize convergence bounds by adaptively compressing model updates and determining the probability of local updates. However, these works predominantly focus on cloud-based FL, which relies on a single cloud server to coordinate the model training.

A few studies~\cite{zhang2024hierarchically,feng2021min,wen2022joint,luo2020hfel,zhou2021two} have explored resource-optimization of HFEL at mobile networks, where multiple edge servers are responsible for aggregating model updates from a subset of devices. In particular, Zhang et al.~\cite{zhang2024hierarchically} proposed a framework for wireless networks, which enhances training efficiency and convergence in multi-cell scenarios by combining intra-cell device-to-device consensus with inter-cell aggregation while optimizing resources to minimize training latency and energy consumption. Feng et al.~\cite{luo2020hfel} proposed to optimize computation and communication resources, alongside edge server associations, to minimize global costs while enhancing training performance in HFEL systems. Zhou et al.~\cite{zhou2021two} propose a joint scheduling and resource allocation scheme to improve convergence rates and reduce energy consumption in the mobile edge. The work closest to us is~\cite{wu2023joint}, which considers the cost efficiency of a Multi-Cell FL system. In their setting, the client could establish communication links with multiple servers. Their focus is client association and coordinator node selection. Different from these studies, our goal in this paper is to minimize the total learning latency while preserving the model accuracy by jointly optimizing the topology for edge servers, as well as the communication and computation resource allocations for edge devices, under both system and data heterogeneity. 



\section{Preliminaries and Problem Formulation}\label{sec:sys_model}
\begin{figure}[t]
\centering
\includegraphics[width=0.48\textwidth]{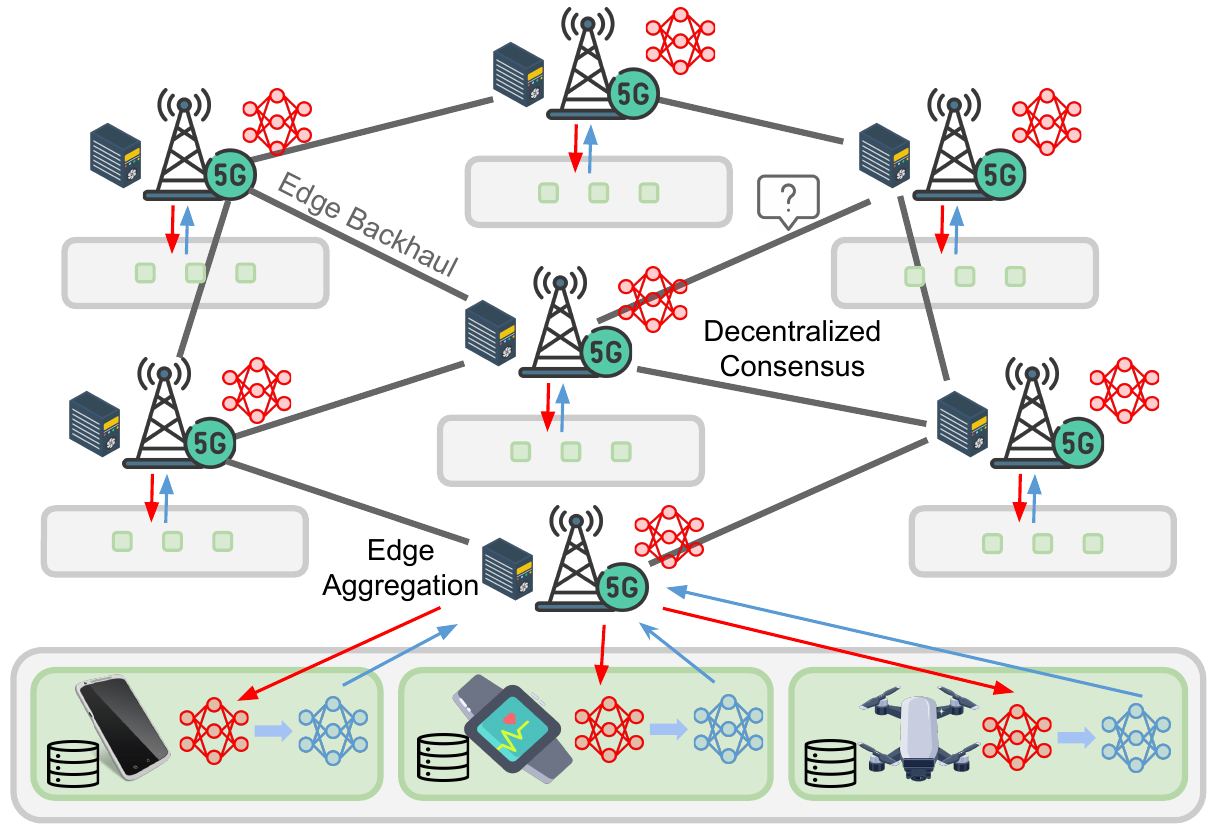}
\caption{Flow-chart of a two-tier federated learning system.}\label{fig:system}
\end{figure}

\subsection{Federated Learning over Mobile Edge}
We consider a hierarchical federated learning system as shown in Fig.~\ref{fig:system}. Assume there are $C$ clusters in the system. Every cluster $c \in [C]$ possesses a server colocated with a base station. Each cluster owns a set of devices $\mathcal{S}_c$, and the number of devices is $N_c = \lvert \mathcal{S}_c\rvert$. Note the devices in $\mathcal{S}_c$ only communicate with the server within the same cluster via the wireless communication links, e.g., 5G. We define the set of all devices in the system as $\mathcal{S} = \cup_{c = 1}^{C} \mathcal{S}_c$. The total number of devices is $N = \lvert \mathcal{S}\rvert$. 


The edge backhaul communication pattern is defined as $\mathcal{G}_{b} = \{\mathcal{V}, \mathcal{E}_{b}\}$, which is an undirected and connected graph. Here $\mathcal{V}$ denotes the set of all servers, and $\mathcal{E}_{b}$ is the set of possible communication links. Moreover, let $\mathbf{A}_{b}$ be the adjacency matrix of $\mathcal{G}_{b}$, where $(\mathbf{A}_{b})_{c,c^{\prime}} = 1$ if there is an edge between server $c$ and server $c^{\prime}$ and $(\mathbf{A}_{b})_{c,c^{\prime}} = 0$ otherwise.

\subsection{Model Training Process}
The objective of FL is to find a model $\mathbf{w} \in \mathbb{R}^d$ that minimize the following global objective function: 
\begin{equation}\label{prob:overall}
\min_{\mathbf{w}} F(\mathbf{w}) = \frac{1}{N}\sum_{n=1}^{N} F_n(\mathbf{w}),
\end{equation}
where $F_n(\mathbf{w}) = \mathbb{E}_{x \sim \mathcal{D}_n}[\ell_n(\mathbf{w}; x)]$ is the local objective function of device $n$, $\mathcal{D}_n$ is the data distribution of device $n$. Here $\ell_n$ is the loss function, e.g., cross-entropy, and $x$ denotes a data sample drawn from the distribution $\mathcal{D}_n$. We define the cluster-level objective function as
\begin{equation}\label{prob:edge}
\min_{\mathbf{w}} f_c(\mathbf{w}) = \frac{1}{N_c}\sum_{n \in \mathcal{S}_c}  F_n(\mathbf{w}),
\end{equation}
Here, $f_c$ is the objective function of the $c$-th cluster, which is the average of the local objective of all devices from cluster $c$. Then we can rewrite the global objective function~\eqref{prob:overall} as:
\begin{equation}\label{prob:global_edge}
\min_{\mathbf{w}} F(\mathbf{w}) = \sum_{c=1}^C \frac{N_c}{N} f_c(\mathbf{w}).
\end{equation}

We summarize the detailed training process in \textbf{Algorithm \ref{algorithm-1}}. Specifically, we assume there is a central controller that will coordinate the operations of the HFEL system. It collects the device-specific parameters, including the CPU cycle per sample $c_n$, the size of the training data $D_n$, energy budget $\bar{\mathcal{E}}_n$, the capacitance coefficient $\alpha_n$, decision boundaries $f_{n}^{\text{min}}$, $f_{n}^{\text{max}}$, $p_{n}^{\text{min}}$, $p_{n}^{\text{max}}$, and background noise power $N_0$, before the training starts. The extra costs (e.g., bandwidth consumption and time cost) for information collection are ignored as in prior works~\cite{luo2022tackling,tran2019federated,luo2021cost,perazzone2022communication,li2023anycostfl,wu2023joint,wang2023federated}. Other inputs are hyper-parameters, e.g., sampling frequency $K$, batch size $I$, and local iterations $S$ depending on user specification. 

At the beginning of $t$-th global round, each server collects its real-time communication bandwidth to other servers $\mathbf{B}^t$ and sends it to the coordinator (line 2). Each global round includes $R$ edge rounds. At the beginning of $r$-th edge round, each edge device records its observed signal-to-noise ratio $\text{SNR}_n^{t,r}$ and communication bandwidth to connected edge server $b_{c'}^{t,r}$, and then sends them to the coordinator (line 5). Coordinator determines the allocated bandwidth for each client $b_n^{t,r}$, CPU frequency $f_n^{t,r}$, and graph topology $\mathcal{G}^t$ by solving \textbf{Algorithm 2} (line 6). The solving details will be elaborated in Section~\ref{sec:problem_solve}. After that, server $c$ broadcasts the latest global model $\bf{u}_c^{t,r}$ to all devices in $\mathcal{S}_c$ (line 7). Then, devices receive $\bf{u}_c^{t,r}$ and initialize their local model to be the received global model (line 9). Next, the device runs $S$ iterations local update (SGD) on its local training dataset (lines 10-13). After local training, the device uploads the model to the server $c$ (line 14). Server $c$ aggregates the local models from devices in $\mathcal{S}_c$ (line 16). The edge training lasts for $R$ rounds. After edge training, server $c$ communicates $\psi$ times with its neighbor server $c' \in \mathcal{N}_c$ to synchronize the global model (lines 18-19).

\begin{algorithm}[t]
\textbf{Input:} Initial global model $\mathbf{u}_{c}^{0,0}$, learning rate $\eta$, mixing matrix $M$, $T, R, S$, $\psi$\\
\textbf{Output:} Final learned model $\mathbf{u}^{T-1}$

\caption{Training Process of FL over Mobile Edge}\label{algorithm-1}
\begin{algorithmic}[1]
    \FOR{each global round $t = 0, \ldots, T-1$}
        \FOR{each cluster $c \in [C]$ \textbf{in parallel}}
            \FOR{each edge round $r = 0, \ldots, R-1$}
                \STATE Server $c$ broadcasts $\mathbf{u}^{t, r}_{c}$ to all devices in $\mathcal{S}_{c}$\label{ln1:1}
                \FOR{Device $n \in \mathcal{S}_c$ \textbf{in parallel}}
                    \STATE $\mathbf{w}^{t, r, 0}_{n} \gets \mathbf{u}^{t, r}_{c}$ \label{ln1:2}
                    \FOR{$s = 0, \ldots, S-1$}\label{ln1:2.5}
                        \STATE Compute a stochastic gradient $\mathbf{g}_n$ over a mini-batch $\xi_n$ drawn from $\mathcal{D}_n$ \label{ln1:3}
                        \STATE $\mathbf{w}^{t, r, s + 1}_{n} \gets \mathbf{w}^{t, r, s}_{n} - \eta \mathbf{g}_n(\mathbf{w}^{t, r, s}_{n})$\label{ln1:4}
                    \ENDFOR\label{ln1:4.5}
                    \STATE Upload $\mathbf{w}^{t, r, S}_{n}$ to server $c$\label{ln1:5}
                \ENDFOR
             \STATE $\mathbf{u}^{t, r + 1}_{c} \gets \frac{1}{N_c}\sum_{n \in \mathcal{S}_c}\mathbf{w}^{t, r, S}_{n}$ \label{ln1:6}
            \ENDFOR
            \STATE {\small $\mathbf{u}^{t + 1, 0}_{c} \gets \mathbf{u}^{t, R-1}_{c} + \displaystyle 
 \sum_{c^{\prime} \in \{c\}\cup\mathcal{N}_c^{t}}\mathbf{M}^{\psi}_{c, c^{\prime}} (\mathbf{u}^{t, R-1}_{c^{\prime}} -\mathbf{u}^{t, R-1}_{c})$ } \label{ln1:7}
        \ENDFOR
    \ENDFOR
\end{algorithmic}
\end{algorithm}

\subsection{Cost Analysis of HFEL}


\subsubsection{Device Communication Time}
We assume the communication follows the Frequency Division Multiple Access (FDMA) protocol, and each server adaptively assigns its communication bandwidth to all devices within the cluster. Let $b_{n}^{t,r}$ be the assigned bandwidth for device $n$ at edge round $r$ and global round $t$. Then, the communication rate of device $n$ is
\begin{equation}\label{equ:com_rate}
\begin{split}
r_{n}^{t,r} = b_{n}^{t,r}\log_{2}(1+\text{SNR}_{n}^{t,r}  ),
\end{split}
\end{equation}
Here, $\text{SNR}_{n}^{t,r}$ denotes the signal-to-noise ratio between device $n$ and the corresponding server at edge round $r$ and global round $t$. 

The totally assigned bandwidth for all devices in $\mathcal{S}_c$ should not exceed the available bandwidth of server $c$, therefore we have
\begin{equation}\label{equ:cons_bu}
    \begin{split}
        \sum_{n \in \mathcal{S}_{c}} b_{n}^{t,r} \leq b_{c}^{t,r}, \forall c,\forall t \\
        b_{n}^{t,r}  > 0, \forall t, \forall r, \forall n,
    \end{split}
\end{equation}
where $b_{c}^{t,r}$ denotes the available bandwidth of the server $c$ at edge round $r$ and global round $t$. The communication time $\mathcal{T}_{n}^{t,r,\text{com}}$ of device $n$ can be expressed as
\begin{equation}\label{equ:com_time}
\begin{split}
\mathcal{T}_{n}^{t,r,\text{com}} &=\frac{\Lambda}{r_{n}^{t,r}}= \frac{\Lambda }{b_{n}^{t,r} \log_{2}(1+\text{SNR}_{n}^{r,t})}.
\end{split}
\end{equation}
Here, $\Lambda$ denotes the model size. Note that we only consider the upload time cost since the download time cost is not the bottleneck for the practical HFEL system.

 



\subsubsection{Server Communication Time}
Define $\mathbf{B}^{t}\in \mathbb{R}^{C \times C}$ as the matrix of communication bandwidth between servers at global round $t$, where $\mathbf{B}_{c,c^{\prime}}^{t}$ is the available bandwidth between server $c$ and $c^{\prime}$. If there is no available link between $c$ and $c^{\prime}$, we set $\mathbf{B}_{c,c^{\prime}}^{t}=0, \forall (c,c^{\prime}) \notin \mathcal{E}_b$. It is worth noting that the communication between servers is accomplished through the edge backhaul, which operates independently from the communication between the server and devices (e.g., through the base station). 

In this paper, we consider the topology design approach that adaptively selects and deletes the slow communication links from the base graph $\mathcal{G}_b$. Specifically, we aim to find an edge backhaul communication pattern $\mathcal{G}^{t} = \{V, E^{t}\}$ at each global round. Here $E^{t} \subset E_{b}$ denotes the set of remaining edges in $\mathcal{G}^{t}$. Note different choices of $\mathcal{G}^{t}$ have an impact on model convergence. We will discuss it in a later section. Then, the communication time between server $c$ and its neighbors depends on the slowest links, which can be expressed as
\begin{equation}
    \mathcal{T}_{c}^{t} = \frac{\psi \Lambda}{ \underset{c^{\prime} \in \mathcal{N}_{c,}^{t}  } { \text{min} } \{\mathbf{B}_{c,c^{\prime}}^{t} \} }.
\end{equation}
Here $\mathcal{N}_c^t$ is the set of neighbors for server $c$ in $\mathcal{G}^{t}$.

Let $\mathbf{A}^t$ be the adjacency matrix of $\mathcal{G}^t$, and $\mathbf{D}^t$ be the degree matrix. Here $\mathbf{D}^t$ is a diagonal matrix, and each element denotes the number of neighbors $\mathbf{D}^t_{c,c} = |\mathcal{N}_{c}^t|$. Then, the Laplacian matrix $\mathbf{L}^t$ of $\mathcal{G}^t$ can be expressed as 
\begin{equation}\label{equ:graph_conn}
    \mathbf{L}^t = \mathbf{D}^t - \mathbf{A}^t.
\end{equation}
We always require the graph $\mathcal{G}^t$  to remain connected. According to the spectral graph theory~\cite{chung1997spectral}, it can be translate into following constraint:
\begin{equation}\label{equ:time_cp}
    \lambda_{2}(\mathbf{L}^t)>0, \forall t
\end{equation}
where $\lambda_{l}(\mathbf{L}^t)$ denotes the $l$-th smallest eigenvalue of Laplacian matrix $\mathbf{L}^t$. 

\subsubsection{Edge Device Computation Time}
Let $\mu_n$ represent the number of CPU cycles required by device $n$ to process a single training example. The value of $\mu_n$ can either be measured offline or known as a prior. The total number of CPU cycles required to train one edge round is $S I \mu_n$, where $I$ denotes the batch size. The computation time for one edge round of device $n$ can be formulated as
\begin{equation}
    \mathcal{T}_{n}^{t,r,\text{cmp}} = \frac{S I \mu_n}{f_{n}^{t,r}}, 
\end{equation}
where $f_{n}^{t,r}$ is the CPU frequency of device $n$ at edge round $r$ and global round $t$. 

\subsubsection{Time Model}
In Algorithm~\ref{algorithm-1}, one global round comprises $R$ edge rounds conducted within the cluster, followed by $\psi$ times synchronization between the clusters. The time required for one edge round depends on the slowest device in that round. Therefore, we have
\begin{equation}
    \mathcal{T}_{c}^{t,r} = \underset{n \in \mathcal{S}_{c} } { \text{max} } \{\mathcal{T}_{n}^{t,r,\text{cmp}} + \mathcal{T}_{n}^{t,r,\text{com}}\},
\end{equation}
where $\mathcal{T}_{c}^{t,r}$ is the time cost of cluster $c$ at edge round $r$ and global round $t$.

After $R$ edge rounds training within the cluster, the server starts to communicate with neighbor servers. Similarly, the slowest cluster determines the final completion time for one global round
\begin{align}
    \mathcal{T}^{t} = \underset{c \in [ C ] } { \text{max} } \{ \textstyle\sum_{r=0}^{R-1} \underset{n \in \mathcal{S}_{c} } { \text{max} } \{\mathcal{T}_{n}^{t,r,\text{cmp}} + \mathcal{T}_{n}^{t,r,\text{com}}\} + \mathcal{T}_{c}^{t} \}.
\end{align}
Here, $\mathcal{T}^{t}$ is time consumption at global round $t$.

\subsubsection{Edge Device Computation Energy}
Following~\cite{burd1996processor}, the CPU energy cost for one edge round training can be formulated as
\begin{equation}\label{equ:ene_cp}
    \mathcal{E}_{n}^{t,r,\text{cmp}} = \frac{\alpha_n}{2}  SI \mu_n (f_n^{t,r})^2,
\end{equation}
where $\alpha_n/2$ denotes the effective capacitance coefficient of the computing chipset on device $n$. 

We assume the devices could adjust their CPU frequency by leveraging the Dynamic Voltage and Frequency Scaling (DVFS) technique. Due to the hardware limitation, the CPU frequency satisfy
\begin{equation}\label{equ:f_const}
    f_{n}^{\text{min}} \leq f_n^t \leq f_n^{\text{max}}
\end{equation}
Here $f_{n}^{\text{min}}, f_n^{\text{max}}$ denote the minimum and maximum CPU frequency of device $n$. 


\subsubsection{Edge Device Communication Energy}
For devices, the energy used for downloading is usually negligible. Therefore, we only consider the communication energy usage during uploading
\begin{equation}
\begin{split}
    \mathcal{E}_{n}^{t,r,\text{com}} =p_n \mathcal{T}_{n}^{t,r,\text{com}}= \frac{p_n \Lambda }{b_{n}^{t,r} \log_{2}(1+\text{SNR}_{n}^{r,t})},
\end{split}
\end{equation}
where $p_n$ denotes the communication power of device $n$.

\subsubsection{Energy Model}
As the server is usually equipped with a plug-in power supply, the energy cost of the server is not our focus in this paper. For devices, the energy cost has two sources: the energy used for local training and the energy used for wireless transmission between the devices and server. Thus, we have 
\begin{equation}
\begin{split}
    \mathcal{E}_{n}^{t,r} =\frac{p_n\Lambda }{b_{n}^{t,r} \log_{2}(1+\text{SNR}_{n}^{r,t})} + \frac{\alpha_n}{2}SI\mu_{n}(f_{n}^{t,r})^{2},
\end{split}
\end{equation}
where $\mathcal{E}_{n}^{t,r}$ is the energy consumption of device $n$ at edge round $r$ and global round $t$.

\subsection{Inter-Cluster Gradient Divergence}
In Algorithm~\ref{algorithm-1}, there is no actual global model, and each server hosts an edge model that serves as the ``global model'' within the cluster. The edge models belonging to different clusters are usually not the same. Therefore, we introduce the consensus distance to measure the discrepancy between any two edge models
\begin{align}\label{equ:consensus_define}
    \Upsilon_{c,c^{\prime}}^{t} = \|\mathbf{u}^{t, R-1}_{c} - \mathbf{u}^{t, R-1}_{c^{\prime}}\|.
\end{align}
Here $\Upsilon_{c,c^{\prime}}^{t}$ denotes the consensus distance between the edge model $c$ and the edge model $c^{\prime}$ at global round $t$. 

We define the consensus distance between the edge model and the actual global model
\begin{align}
    \Upsilon_{c}^{t} = \|\mathbf{\bar{u}}^{t,0} - \mathbf{u}^{t, 0}_{c}\|,
\end{align}
where $\mathbf{\bar{u}}^{t,0} = \frac{1}{C}\sum_{c=1}^{C}\mathbf{u}^{t, 0}_{c}$ denotes the average of all edge models. Note $\mathbf{\bar{u}}^{t, 0}$ is not available. Moreover, the average consensus distance of all edge models is 
\begin{equation}
    \textstyle \Upsilon^{t} = \frac{1}{C}\sum_{c=1}^{C}\Upsilon_{c}^{t}.
\end{equation}
Similar to the weight divergence and consensus distance in prior works\cite{liao2022adaptive,wang2022accelerating,kong2021consensus}, the consensus distance depends on the data distribution across the clusters, which is the key factor in capturing the joint effect of decentralization. 

\subsection{Problem Formulation and Challenges}

\subsubsection{Formulated Optimization Problem}
Our goal is to minimize the total time cost (during local updating and model transmission) while ensuring model convergence. Specifically, we aim to devise a control strategy by adjusting the decision variables, e.g., graph topology, communication bandwidth, and CPU frequency, which translates into the following problem: 
\begin{equation}
\textbf{P1:}\quad \textstyle \min \sum_{t=0}^{T-1}  \mathcal{T}^{t} \notag \\
\end{equation}
\begin{numcases}
     {\text{s.t.}} 
     \Upsilon^{t+1} \leq \Upsilon_{\text{max}}^{t},\forall t \label{p1:3} \\ 
     \textstyle \sum_{t=0}^{T-1}\sum_{r=0}^{R-1}\mathcal{E}_{n}^{t,r} \leq \bar{\mathcal{E}}_{n}, \forall n \label{p1:4} \\
     (\ref{equ:cons_bu}), (\ref{equ:graph_conn}),(\ref{equ:f_const}).\notag
\end{numcases}

The objective of Problem \textbf{P1} represents the total time cost of $T$ global rounds. The inequality (\ref{p1:3}) states the consensus distance should not exceed a threshold $\Upsilon_{\text{max}}^{t}$. The role of this constraint is two-fold. First, we use it to mitigate the negative influence caused by the non-IID data distribution between clusters. Additionally, it ensures the convergence of the model when we update the graph topology.
The inequality (\ref{p1:4}) represents the constraint on the energy consumption, where $\bar{\mathcal{E}}_{n}$ is the energy budget for device $n$. 

\subsubsection{Challenges to address the problem} It is a challenge to solve the Problem \textbf{P1} due to the: (i) Model convergence is influenced by graph topology. If slow links are removed, the convergence rate may slow down, and more resources may be required to finish the training. (ii) Control decisions in the objective function and constraints are time-slot interconnected, yet we lack knowledge of system stats like cluster communication environment and edge backhaul speed. Thus, we need an online algorithm that is not dependent on pre-existing information or system stats assumptions. (iii) The graph topology and resource allocation have different updating frequencies, which prevents online joint optimization.

\section{Convergence Constraint }\label{sec:converge}

\subsection{Convergence Result}
For the convergence of HFEL systems, several studies~\cite{zhang2022scalable,qu2022context,wang2022demystifying} have been conducted. However, these works mainly focus on the convergence analysis of the learning algorithm in HFEL and assume that all edge devices are homogeneous (i.e., their computation, communication, and storage capabilities are the same) and the edge backhaul topology is given a prior. To analyze the impact of topology design on the convergence of proposed algorithms, we re-elaborate the convergence results from prior work~\cite{zhang2022scalable}. 

Let $L$ be the Lipschitz constant, $\varphi = tRS + rS+s$ denote the global iteration index, $\Phi=RTS$ represent the total number of iterations. Additionally, let $\sigma$ be the bound of unbiased stochastic gradient variance, $\epsilon_c$ be the intra-cluster divergence, $\epsilon$ be the inter-cluster divergence, and $\zeta$ be the second largest eigenvalue of the mixing matrix. Furthermore, we define the following constants
\begin{gather}\label{def_Omega12}
\Omega_{1}=\frac{\zeta^{2\psi}}{1-\zeta^{2\psi}}, \quad \Omega_{2}=\frac{1}{1-\zeta^{2\psi}}+\frac{2}{1-\zeta^{\psi}}+\frac{\zeta^{\psi}}{(1-\zeta^{\psi})^{2}},\notag
\end{gather}

Let $\eta \leq \min { \{\frac{1}{2LS}, \frac{1}{2\sqrt{2\Omega_2}LRS} \} }$, we have
\begin{align}\label{eq:convergence}
\frac{1}{\Phi}\sum_{\varphi=0}^{\Phi-1}&\mathbb{E}\|\nabla F(\mathbf{u}^{\varphi})\|^2  \leq   \frac{2(F(\mathbf{u}^1)-F_{\inf})}{\eta \Phi}+\frac{\eta L\sigma^2}{N} \nonumber \\
&+ 8\eta^2L^2(\Omega_1RS+\frac{C-1}{N}RS)\sigma^2+16\eta^2L^2R^2S^2\Omega_2 \epsilon^2 \nonumber \\
&+8\frac{N-C}{N}\eta^2L^2S \sigma^2+16L^2\eta^2S^2  \sum_{c=1}^C\frac{N_c}{N}\epsilon_c^2.\notag
\end{align}

The term $16\eta^2L^2R^2S^2\Omega_2 \epsilon^2$ captures the inter-cluster error bound. A more sparse graph topology results in a large value of $\Omega_2$~\cite{zhang2022scalable}, thereby increasing the adverse impact of inter-cluster error. This observation necessitates considering topology in model convergence when designing the HFEL system. For instance, if an edge model has a larger consensus distance, encouraging it to connect with neighboring servers more frequently can be beneficial for algorithm convergence and the performance of the learned model. On the other hand, if an edge model has a smaller consensus distance, we can safely remove communication links to other servers, thereby reducing training latency.

\subsection{Consensus Distance Estimation}
Here, we introduce the approach that estimates the consensus distance between the server model $\mathbf{u}^{t,R-1}_{c}$ and the global averaged model $\mathbf{\bar{u}}^{t,R-1}$. Our goal is to capture the relationship between the graph topology and the consensus distance. Based on the definition of consensus distance~(\ref{equ:consensus_define}) and the aggregation rule~(\ref{ln1:7}), we have
\begin{align}
    & \Upsilon_{c}^{t+1} = \big\|\mathbf{\bar{u}}^{t+1,0} - \mathbf{u}^{t+1,0}_{c}\big\| \notag \\
    & \scalebox{0.92}{$=\big\| \displaystyle \frac{1}{C}\displaystyle\sum_{c^{\prime}=1}^{C}\mathbf{u}^{t, R-1}_{c^{\prime}} - (\mathbf{u}^{t, R-1}_{c} +  \sum_{c^{\prime} \in \{c\}\cup\mathcal{N}_c^t}\mathbf{M}^{(\psi)}_{c, c^{\prime}} (\mathbf{u}^{t, R-1}_{c^{\prime}} -\mathbf{u}^{t, R-1}_{c})) \big\| $}\notag \\
    & = \big\| \textstyle \sum_{c^{\prime}=1}^{C} \frac{\mathbf{u}^{t, R-1}_{c^{\prime}} - \mathbf{u}^{t, R-1}_{c}}{C} - \mathbf{A}_{c, c^{\prime}} \mathbf{M}^{(\psi)}_{c, c^{\prime}} (\mathbf{u}^{t, R-1}_{c^{\prime}} -\mathbf{u}^{t, R-1}_{c})) \big\|
\end{align}
For simplicity, we set $\mathbf{M}^{(\psi)}_{c, c^{\prime}} = 1/N$. Note it is the possible maximum value~\cite{xiao2004fast,liao2022adaptive,xu2022adaptive}. Then we have
\begin{align}
    \mathbb{E}\Upsilon_{c}^{t+1}  &= \big\|\textstyle\sum_{c^{\prime}=1}^{C} \frac{(1-\mathbf{A}_{c, c^{\prime}})(\mathbf{u}^{t, R-1}_{c^{\prime}} - \mathbf{u}^{t, R-1}_{c})}{C} \big\|\notag \\
    & \leq \textstyle \frac{1}{C} \sum_{c^{\prime}=1}^{C} (1-\mathbf{A}_{c, c^{\prime}})\Upsilon_{c, c^{\prime}}^{t}.\label{equ:consen_expe}
\end{align}
We take the average across all edge models on both sides of~(\ref{equ:consen_expe})
\begin{align}
 \mathbb{E}\Upsilon^{t+1} \leq \frac{1}{C^2} \textstyle\sum_{c=1}^{C}\sum_{c^{\prime}=1}^{C} (1-\mathbf{A}_{c, c^{\prime}})\Upsilon_{c, c^{\prime}}^{t}.\label{equ:consen_upper}
\end{align}

Next, we substitute the upper bound~(\ref{equ:consen_upper}) to~(\ref{p1:3}) and ensure the upper bound is smaller than the threshold. We have the following consensus constraint:
\begin{equation}\label{equ:consensus_new}
    \frac{1}{C^2} \textstyle \sum_{c=1}^{C}\sum_{c^{\prime}=1}^{C} (1-\mathbf{A}_{c, c^{\prime}})\Upsilon_{c, c^{\prime}}^{t} \leq \Upsilon^{t}_{\text{max}}, \forall t.
\end{equation}

The inequality~(\ref{equ:consensus_new}) captures the property of topology design on mitigating the negative impact of inter-cluster non-IID data distribution. Specifically, a relatively large value of $\Upsilon_{c, c^{\prime}}^{t}$ implies a high non-IID degree in data distribution between cluster $c$ and $c^{\prime}$. A relatively smaller value vice verse. A larger $\Upsilon_{c, c^{\prime}}^{t}$ suggests a substantial disparity of the edge models on two servers, underscoring the importance of maintaining collaborative training between them. Consequently, disconnecting the communication link between two servers may degrade the overall convergence rate. Conversely, a smaller consensus distance indicates that the differences in data distribution are minimal. In this scenario, the communication link between the servers can be safely removed, thereby reducing the communication overhead among the edge servers. Note that if two servers are not connected, directly computing $\Upsilon_{c,c^{\prime}}^{t}$ is infeasible. In this case, we adopt the approach in prior works [34], [35] to estimate $\Upsilon_{c,c^{\prime}}^{t}$.

\section{Solution Design}\label{sec:problem_solve}

\begin{algorithm}[t]
\caption{FedRT}\label{alg:FedRT}
\begin{algorithmic}[1]
    \FOR{$t=0$ to $T-1$}
        \FOR{$r=0$ to $R-1$}
            \IF{$r=R-1$}
                \STATE Server records $\mathbf{B}_{c,}$ and $\Upsilon_{c,c^{\prime}}^{t}$, and send them to the coordinator (it can be any server);
            \ENDIF
            \STATE Device observes channel statistics $\text{SNR}_{n}^{t,r}$ and send it to coordinator;
            \IF{$r\neq R-1$}
                \STATE Solve Problem \textbf{P2.1} through CVX;
            \ELSE
                \STATE Solve Problem \textbf{P2.2} via \textbf{Algorithm~\ref{iter}};
            \ENDIF
            
            \STATE Run one edge round collaborative training.
            \IF{$r=R-1$}
                \STATE Server synchronizes model with neighbor servers.
            \ENDIF
        \ENDFOR
    \ENDFOR
\end{algorithmic}
\end{algorithm}
\begin{algorithm}[t]
\textbf{Input:} $\Lambda$, $S$, $I$, $\mathbf{B}^t$, $\zeta$, $\Upsilon_{\text{max}}^{t}$, $\mathcal{G}_{b}$,  $\{\mu_{n}\}$, $\{p_{n}\}$, $\{\alpha_{n}\}$, $\{\text{SNR}_{n}^{t,R}, \forall n\}$, $\{\Upsilon_{c,c^{\prime}}^{t}, \forall c, \forall c^{\prime}\}$ \\
\textbf{Output:} $\mathbf{b}^{t,R-1}, \mathbf{f}^{t,R-1}, \mathcal{G}^{t}$
\caption{Adaptive control algorithm of FedRT}\label{iter}
\begin{algorithmic}[1]
    \STATE Initial topology as $(\mathbf{A}^{t})^{\ast}=\mathbf{A}_{b}$ and $Flag = True$
    \STATE Solve Problem \textbf{P2.2} to obtain an initial resource allocation under base topology 
    

    $(\mathbf{b}^{t,R-1})^{\ast}, (\mathbf{f}^{t,R-1})^{\ast} \leftarrow \argmin\limits_{\mathbf{b}, \mathbf{f}} \mathcal{T}^{t,R-1}(\mathbf{b}, \mathbf{f}, (\mathcal{G}^{t})^{\ast})$
    
    \STATE Compute the minimum completion time using~(\ref{equ:time_compute})
    
    $(\mathcal{T}^{t,R-1})^{\ast} \leftarrow \mathcal{T}^{t,R-1}((\mathbf{b}^{t,R-1})^{\ast}, (\mathbf{f}^{t,R-1})^{\ast}, (\mathcal{G}^{t})^{\ast})$
    \WHILE{$True$}
        \IF{$Flag$}
            \STATE $e= \lfloor \sqrt{\sum_{c,c^{\prime}}(\mathbf{A}_{c,c^{\prime}}^{t})^{\ast}}\rfloor$
        \ELSE
            \STATE $e= \lfloor e/2 \rfloor$
        \ENDIF
        \STATE Select $e$ slowest links into $\bar{\mathcal{E}}$ from $(\mathcal{G}^{t})^{\ast}$ under the consensus distance threshold (\ref{equ:consensus_new});
        \STATE Sort the links in $\bar{\mathcal{E}}$ by speed
        \FOR{$E_{c,c^{\prime}} \in \bar{\mathcal{E}}$}
        \STATE Remove $E_{c,c^{\prime}}$ from $(\mathcal{G}^{t})^{\ast}$
            \IF{$(\mathcal{G}^{t})^{\ast}$ is not connected}
                \STATE Restore $E_{c,c^{\prime}}$ into $(\mathcal{G}^{t})^{\ast}$
            \ENDIF
        \ENDFOR
        \STATE Solve Problem \textbf{P2.2} to obtain new completion time and resource allocation
        
        $(\mathbf{b}^{t,R-1})^{\prime}, (\mathbf{f}^{t,R-1})^{\prime} \leftarrow \argmin\limits_{\mathbf{b}, \mathbf{f}} \mathcal{T}^{t,R-1}(\mathbf{b}, \mathbf{f}, (\mathcal{G}^{t})^{\ast})$

        \STATE $(\mathcal{T}^{t,R-1})^{\prime} \leftarrow \mathcal{T}^{t,R-1}((\mathbf{b}^{t,R-1})^{\prime}, (\mathbf{f}^{t,R-1})^{\prime}, (\mathcal{G}^{t})^{\ast})$
        \IF{$(\mathcal{T}^{t,R-1})^{\prime} < (\mathcal{T}^{t,R-1})^{\ast}$}
        \STATE $(\mathbf{b}^{t,R-1})^{\ast}, (\mathbf{f}^{t,R-1})^{\ast} \leftarrow (\mathbf{b}^{t,R-1})^{\prime}, (\mathbf{f}^{t,R-1})^{\prime}$
        \STATE $(\mathcal{T}^{t,R-1})^{\ast}, (\mathcal{G}^{t,R-1})^{\ast} \leftarrow (\mathcal{T}^{t,R-1})^{\prime}, (\mathcal{G}^{t,R-1})^{\prime}$
        \STATE $Flag \leftarrow True$
        \ELSE
        \STATE $Flag \leftarrow False$
        \ENDIF
        \IF{not $Flag$ and $s == 1$}
        \STATE Break
        \ENDIF
    \ENDWHILE
    \STATE $\mathcal{G}^{t}, \mathbf{b}^{t,R-1}, \mathbf{f}^{t,R-1}\leftarrow (\mathcal{G}^{t})^{\ast}, (\mathbf{b}^{t,R-1})^{\ast}, (\mathbf{f}^{t,R-1})^{\ast}$
\end{algorithmic}
\end{algorithm}



To address Problem \textbf{P1}, we first decompose it into a series of sub-problems alongside the global training round $t$. Then, the sub-problem at global round $t$ is
\begin{gather}
\textbf{P2:}\quad\min_{\{b_{n}^{r,t},\forall n,r\},\{f_{n}^{r,t},\forall n,r\}, \mathcal{G}^{t}} \mathcal{T}^{t}\notag \\
\text{s.t.}\begin{dcases}
      \frac{1}{C^2} \textstyle \sum_{c=1}^{C}\sum_{c^{\prime}=1}^{C} (1-\mathbf{A}^{t}_{c, c^{\prime}})\Upsilon_{c, c^{\prime}}^{t} \leq \Upsilon^{t}_{\text{max}}  \\
     \textstyle \sum_{t^{\prime}=0}^{t-1}\sum_{r=0}^{R-1}\mathcal{E}_{n}^{t^{\prime},r} +(T-t)\sum_{r=0}^{R-1}\mathcal{E}_{n}^{t,r}  \leq \bar{\mathcal{E}}_{n},\forall n \label{equ:energy_t}\\
     (\ref{equ:cons_bu}), (\ref{equ:graph_conn}),(\ref{equ:f_const}).
\end{dcases}\notag
\end{gather}
Here, the second inequality uses the energy cost at the current global round $t$ to estimate the total energy cost of $T$ global rounds.

However, solving the above Problem \textbf{P2} still is infeasible. One obstacle is the server only starts to communicate with neighbor servers after they finish their training task within each cluster. The graph topology should be derived based on the time cost of $R$ edge training rounds and the consensus distance between the server models. At edge round $r < R-1$, the future communication topology is still unknown, which prevents the coordinator from effectively deciding the communication bandwidth and CPU frequency for devices. 

In order to address this challenge, we again break down Problem \textbf{P2} into a sequence of sub-problems across different edge rounds. Specifically, for edge round $r<R-2$, we allocate the CPU frequency and communication bandwidth for each device such that the estimated completion time for each edge round is minimized. At edge round $R-1$, we jointly optimize the CPU frequency, communication bandwidth, and graph topology based on the completion time and the used energy.

Let $\mathbf{b}^{t,r} = \{b_{n}^{t,r}, \forall n \}$, $\mathbf{f}^{t,r} = \{f_{n}^{t,r}, \forall n \}$ for $r= 0,\ldots,R-1$. At global round $t$, the completion time from the first edge round to edge round $r$ can be estimated as follows:
\begin{align}
    \mathcal{T}^{t,r} =  \underset{c \in [ C ] } { \text{max} } 
    \{ & \textstyle \sum_{r^{\prime}=0}^{r-1}\underset{n \in \mathcal{S}_{c} } {\text{max} } \{ \mathcal{T}_{n}^{t,r^{\prime},\text{cmp}}   + \mathcal{T}_{n}^{t,r^{\prime},\text{com}} \} \notag\\
    &\quad+ \underset{n \in \mathcal{S}_{c} } { \text{max} } \{ \mathcal{T}_{n}^{t,r,\text{cmp}} + \mathcal{T}_{n}^{t,r,\text{com}} \} \}
\end{align}
Then, the sub-problem at edge round $r$ and global round $t$ can be expressed as
\begin{gather}
    \textbf{P2.1:}\quad \min_{ \mathbf{b}^{t,r}, \mathbf{f}^{t,r}}   \mathcal{T}^{t,r} \notag \\
     \textrm{s.t.} \begin{dcases}\textstyle\sum_{t^{\prime}=0}^{t-1}\sum_{r^{\prime}=0}^{R-1}\mathcal{E}_{n}^{t^{\prime},r^{\prime}}+\sum_{r^{\prime}=0}^{r-1}\mathcal{E}_{n}^{t,r^{\prime}} \notag\\
     \quad \qquad+((T-t)R+R-r)\mathcal{E}_{n}^{t,r}  \leq \bar{\mathcal{E}}_{n}, \forall n\notag \\
     (\ref{equ:cons_bu}),(\ref{equ:f_const}). \notag
      \end{dcases}
\end{gather}
Note that the problem is convex, and it can be efficiently solved using convex optimization software like CVX.

At edge round $R-1$, the backhaul communication bandwidth matrix $B^{t}$ is available. The completion time of global round $r$ can be estimated as follows:
\begin{equation}
    \begin{split}
        \mathcal{T}^{t,R-1} =  \underset{c \in [ C ] } { \text{max} } 
    \{ &\textstyle\sum_{r^{\prime}=0}^{R-2}\underset{n \in \mathcal{S}_{c} } {\text{max} } \{ \mathcal{T}_{n}^{t,r^{\prime},\text{cmp}}   + \mathcal{T}_{n}^{t,r^{\prime},\text{com}} \}\\
    &\quad+ \underset{n \in \mathcal{S}_{c} } { \text{max} } \{ \mathcal{T}_{n}^{t,R-1,\text{cmp}} + \mathcal{T}_{n}^{t,R-1,\text{com}} \} +\mathcal{G}^{t}\}
    \end{split}\label{equ:time_compute}
\end{equation}
Then, the sub-problem at edge round $R-1$ and global round $t$ can be formulated as
\begin{gather}
    \textbf{P2.2:}\quad \textstyle \min_{ \mathbf{b}^{t,R-1}, \mathbf{f}^{t,R-1},\mathcal{G}^{t}}  \mathcal{T}^{t,R-1} \notag \\
     \textrm{s.t.}
     \begin{dcases}
      \frac{1}{C^2} \textstyle \sum_{c=1}^{C}\sum_{c^{\prime}=1}^{C} (1-\mathbf{A}^{t}_{c, c^{\prime}})\Upsilon_{c, c^{\prime}}^{t} \leq \Upsilon^{t}_{\text{max}} \\
     \textstyle \sum_{t^{\prime}=0}^{t-1}\sum_{r^{\prime}=0}^{R-1}\mathcal{E}_{n}^{t^{\prime},r^{\prime}} +\sum_{r^{\prime}=0}^{R-2}\mathcal{E}_{n}^{t,r^{\prime}}\\
     \quad+((T-t)R+1)\mathcal{E}_{n}^{t,R-1} \leq \bar{\mathcal{E}}_{n}, \forall n\notag \\(\ref{equ:cons_bu}),(\ref{equ:graph_conn}),(\ref{equ:f_const}). \notag
    \end{dcases}
\end{gather}

The Problem \textbf{P2.2} can be solved using greedy search. Specifically, we alternatively update the allocated communication bandwidth $\mathbf{b}^{t,R-1}$ and CPU frequency $\mathbf{f}^{t,R-1}$ for each device, and the server communication graph topology $\mathcal{G}^t$ at last edge round until the completion time can not be reduced anymore. Subsequently, the entire problem can be effectively solved online.

We summarize the overall solving process in~\textbf{Algorithm~\ref{alg:FedRT}}. Specifically, at the beginning of each edge round, each server records its real-time communication bandwidth with other servers and sends it to the coordinator. Then, the server computes the consensus distance to the other server’s model using the latest aggregated model. The consensus distance will be uploaded to the coordinator to update the consensus distance matrix between the servers. Next, each device collects channel statistics $\text{SNR}_n^{t,r}$ and sends it to the coordinator for problem-solving (line 6). If the current edge round is not the last edge round, we directly solve the Problem P2.1 using convex optimization software like CVX to obtain $\mathbf{b}^{t,r}$ and $f^{t,r}$ (line 8). If the current edge round is the last round, we will jointly choose the $\mathbf{b}^{t,R-1}$, $\mathbf{f}^{t,R-1}$, and $\mathcal{G}^t$ by solving Problem P2.2 (line 10). After the $R$ edge training round, the server will communicate with connected neighbor servers in $\mathcal{G}^t$ to achieve global consensus (line 14).

The process for solving Problem \textbf{P2.2} is summarized in~\textbf{Algorithm~\ref{iter}}. Specifically, the coordinator first initializes the graph topology $(\mathbf{A}^t)^*$ as the base graph topology $\mathbf{A}_b$ and utilizes it to obtain the initial bandwidth allocation $(\mathbf{b}^{t,R-1})^*$ and frequency assignment $(\mathbf{f}^{t,R-1})^*$ (line 1-2). The initial complete time $(\mathcal{T}^{t,R-1})^*$ is then computed based on obtained $(\mathbf{b}^{t,R-1})^*$, $(\mathbf{f}^{t,R-1})^*$, $(\mathcal{G}^t)^*$ (line 3). Next, we will alternatively update the $\mathbf{b}^{t,R-1}$ and $\mathbf{f}^{t,R-1}$ with $\mathcal{G}^t$ until the completion time can not be reduced anymore. Our method gradually removes the slowest link from the base graph topology. In particular, we first sort all links in the initial graph topology (line 10), and we then select $e$ slowest links from the graph without compromising the consensus distance constraint (26) (line 11). For each selected link, we gradually remove it from the initial graph topology while keeping the graph connected (lines 12-17). After that, we solve the Problem P2.2 using the obtained graph topology to solve $\mathbf{b}^{t,R-1}$ and $\mathbf{f}^{t,R-1}$ (line 18). The new completion time is then updated using the obtained $\mathbf{b}^{t,R-1}$, $\mathbf{f}^{t,R-1}$ and $\mathcal{G}^t$. If the new completion time is smaller than the previous completion time, we will continue the next search round. If the new completion time is larger than the previous completion time, we will reduce the number of removed links $e$ by half (line 8). We set the initial number of removed links to be the square of the number of all links in the base graph (line 6). The search processing will be terminated until there is a link that can be removed from the graph (line 27). Finally, the obtained $\mathbf{b}^{t,R-1}$, $\mathbf{f}^{t,R-1}$ and $\mathcal{G}^t$ will be returned (line 30).

\section{Experiment}\label{sec:exp}
\begin{figure}[t]
    \centering
    \subfloat[]{{\includegraphics[width=0.2\textwidth]{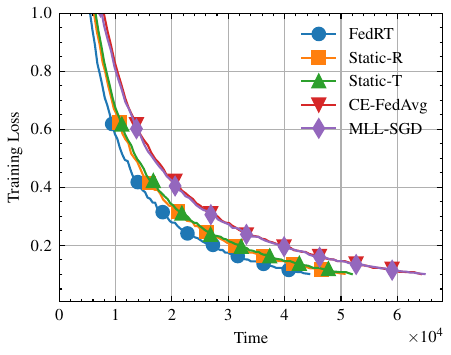}}} \hspace{0.4cm}
    \subfloat[]{{\includegraphics[width=0.2\textwidth]{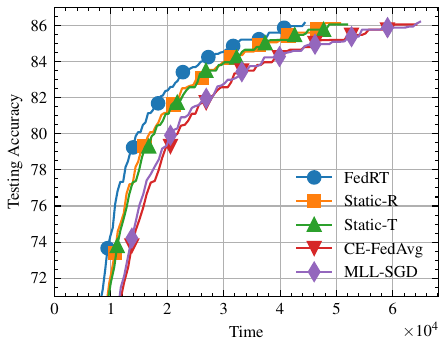}}} \\
    \subfloat[]{{\includegraphics[width=0.2\textwidth]{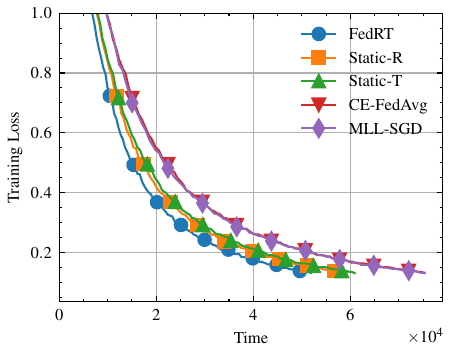}}} \hspace{0.4cm}
    \subfloat[]{{\includegraphics[width=0.2\textwidth]{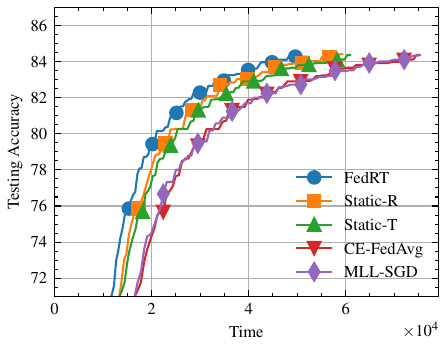}}} 
    \caption{Training loss (a,c) and testing accuracy (b,d) comparison of FedRT and baselines under IID (a,b) and non-IID (c,d) distribution on CIFAR-10.}
    \label{fig:baseline_cifar}%
\end{figure}

\begin{figure}[t]
    \centering
    \subfloat[]{{\includegraphics[width=0.2\textwidth]{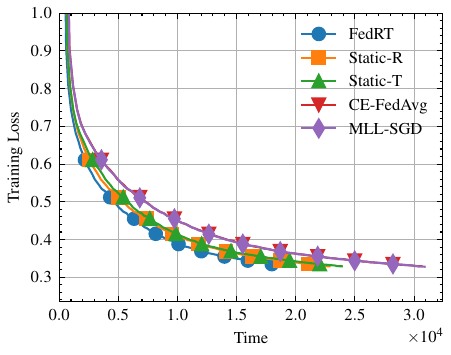}}} \hspace{0.4cm}
    \subfloat[]{{\includegraphics[width=0.2\textwidth]{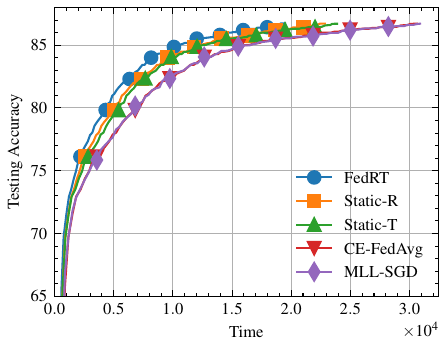}}} \\
    \subfloat[]{{\includegraphics[width=0.2\textwidth]{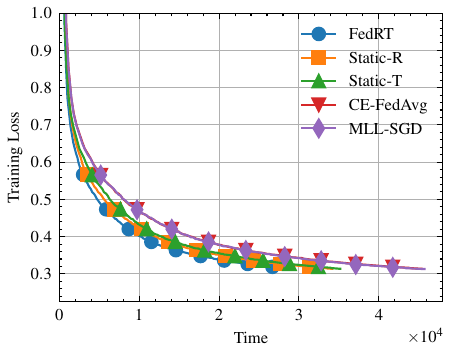}}} \hspace{0.4cm}
    \subfloat[]{{\includegraphics[width=0.2\textwidth]{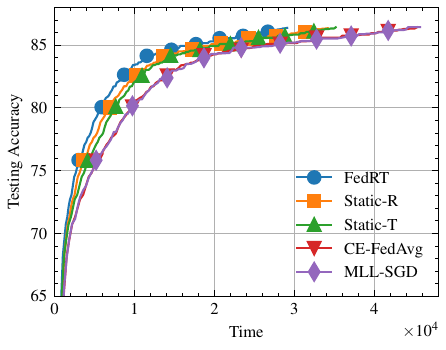}}} 
    \caption{Training loss (a,c) and testing accuracy (b,d) comparison of FedRT and baselines under IID (a,b) and non-IID (c,d) distribution on FMNIST.}
    \label{fig:baseline_fmnist}%
\end{figure}

\begin{figure}[t]
    \centering
    \subfloat[]{{\includegraphics[width=0.2\textwidth]{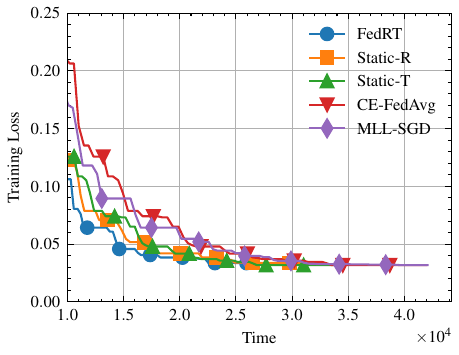}}} \hspace{0.4cm}
    \subfloat[]{{\includegraphics[width=0.2\textwidth]{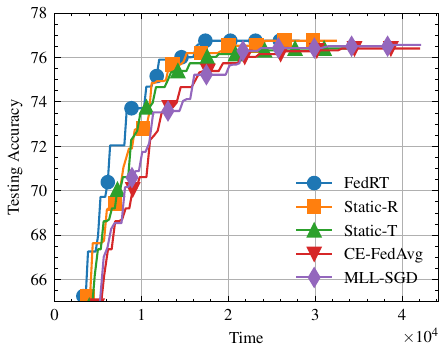}}} 
    \caption{Training loss (a,c) and testing accuracy (b,d) comparison of FedRT and baselines under IID (a,b) and non-IID (c,d) distribution on FEMNIST.}
    \label{fig:baseline_femnist}%
\end{figure}

\begin{figure}[t]
    \centering
    \subfloat[]{{\includegraphics[width=0.2\textwidth]{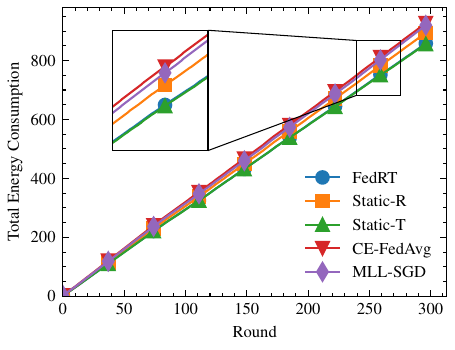}}} \hspace{0.4cm}
    \subfloat[]{{\includegraphics[width=0.2\textwidth]{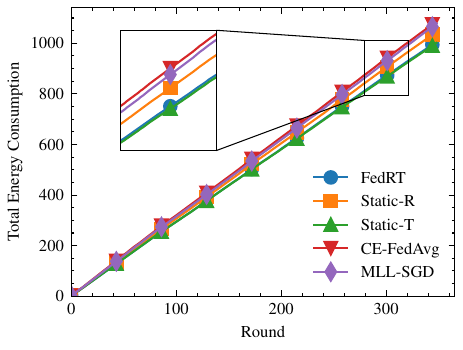}}} 
    \caption{Cumulative energy consumption comparison of FedRT and baselines under IID (a) and non-IID (b) distribution on CIFAR-10.}
    \label{fig:ene_cifar}%
\end{figure}

\begin{figure}[t]
    \centering
    \subfloat[]{{\includegraphics[width=0.2\textwidth]{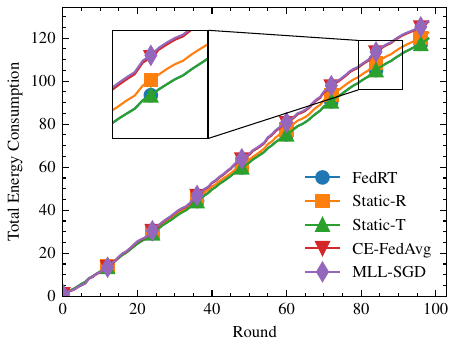}}} \hspace{0.4cm}
    \subfloat[]{{\includegraphics[width=0.2\textwidth]{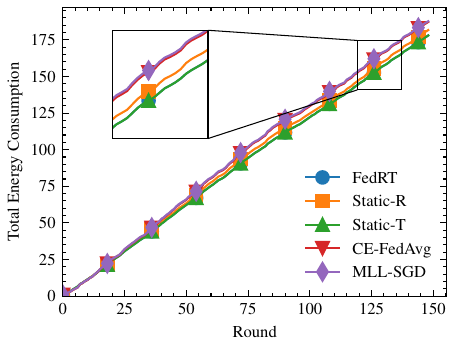}}} 
    \caption{Cumulative energy consumption comparison of FedRT and baselines under IID (a) and non-IID (b) distribution on FMNIST.}
    \label{fig:ene_fmnist}%
\end{figure}

\begin{figure}[t]
    \centering
    \subfloat[]{{\includegraphics[width=0.2\textwidth]{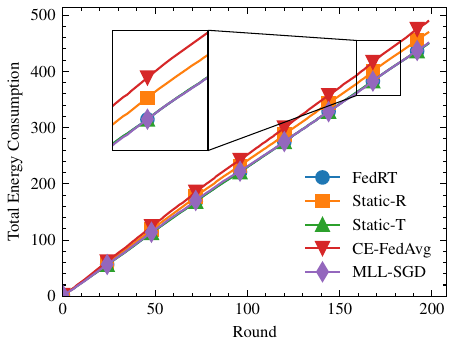}}} 
    \caption{Cumulative energy consumption comparison of FedRT and baselines on FEMNIST.}
    \label{fig:ene_femnist}%
\end{figure}


\begin{figure}[t]
    \centering\subfloat[CIFAR-10]{{\includegraphics[width=0.2\textwidth]{ {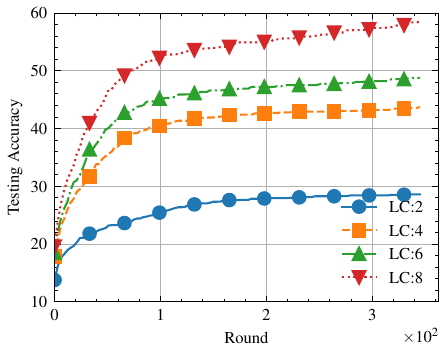}} }} \hspace{0.4cm}
    \centering\subfloat[FMNIST]{{\includegraphics[width=0.2\textwidth]{ {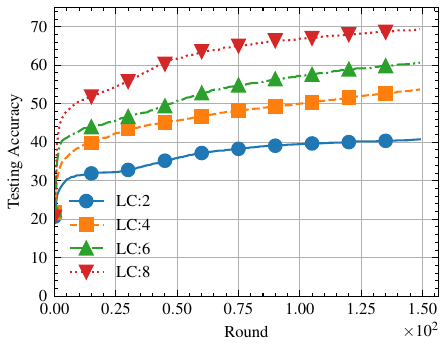}} }} 
    \caption{FedRT under different levels of inter-cluster data heterogeneity.}
    \label{fig:lc_noniid}%
\end{figure}

\begin{figure}[t]
    \centering\subfloat[CIFAR-10]{{\includegraphics[width=0.2\textwidth]{ {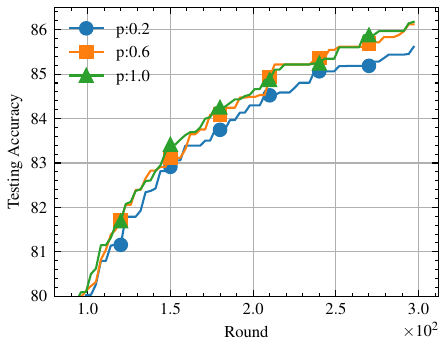}} }} \hspace{0.4cm}
    \centering\subfloat[FEMNIST]{{\includegraphics[width=0.2\textwidth]{ {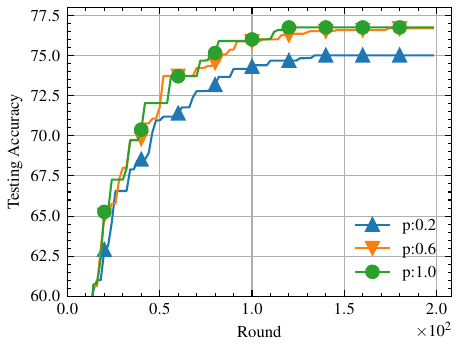}} }} 
    \caption{The performance of FedRT under different base graph topologies.}
    \label{fig:topo}%
\end{figure}

\begin{table*}[t]
  \caption{Test accuracy and resource usage comparison of FedRT and baselines for different datasets. non-IID is Dirichlet distribution. Time denotes the total training latency in hours. Eng represents the total energy consumption (J). Acc is the best testing accuracy. }
  \label{compare-table}
  \centering
  \resizebox{2\columnwidth}{!}{
    \begin{tabular}{ccccccccccccccccc}
    \midrule
    \multirow{4}{*}{\textbf{Method} } & \multicolumn{6}{c}{\textbf{CIFAR-10} } & \multicolumn{6}{c}{\textbf{FMNIST}} & \multicolumn{3}{c}{\textbf{FEMNIST}} \\ \cmidrule(lr){2-7} \cmidrule(lr){8-13} \cmidrule(lr){14-16}
      & \multicolumn{3}{c}{IID }  & \multicolumn{3}{c}{non-IID } & \multicolumn{3}{c}{IID }  & \multicolumn{3}{c}{non-IID } & \multicolumn{3}{c}{/} \\ \cmidrule(lr){2-4} \cmidrule(lr){5-7} \cmidrule(lr){8-10} \cmidrule(lr){11-13} \cmidrule(lr){14-16}
      & Time \(\downarrow\)  & Eng \(\downarrow\) & Acc \(\uparrow\) & Time \(\downarrow\)  & Eng \(\downarrow\) & Acc \(\uparrow\) & Time \(\downarrow\)  & Eng \(\downarrow\) & Acc \(\uparrow\)& Time \(\downarrow\)  & Eng \(\downarrow\) & Acc \(\uparrow\) & Time \(\downarrow\)  & Eng \(\downarrow\) & Acc \(\uparrow\) \\
    \hline
  FedRT & \textbf{12.41}  & 864.12 & \textbf{86.14} & \textbf{14.46} & 1004.05 & 84.28 &  \textbf{5.41} & \textbf{119.81} & \textbf{86.73} & \textbf{8.08} & 177.93 & 86.38 & \textbf{7.86} & 450.55& \textbf{76.76}\\
 Static-R & 14.16  & 899.06 & 86.13 & 16.52 & 1047.93 & \textbf{84.42} &  6.35 & 122.52 & 86.72 & 9.49 & 181.57 & 86.35 & 9.04 & 470.23 & \textbf{76.76}\\
 Static-T & 14.52  & \textbf{861.47} & 86.06 & 16.98 & \textbf{1001.58} & 84.37 &  6.63 & 119.83 & 86.70 & 9.77 & \textbf{177.92} & 86.42 & 9.40 & \textbf{449.51} & 76.41\\
 CE-FedAvg & 17.95  & 933.47 & 86.06 & 20.99 & 1087.29 & 84.37 &  8.51 & 127.57 & 86.70 & 12.59& 187.10 & 86.42 & 11.73 & 489.42 & 76.41\\
 MLL-SGD & 17.96  & 925.99 & 86.10 & 21.00 & 1078.44 & 84.33 &  8.51 & 127.93 & 86.68 & 12.59& 187.65 & \textbf{86.44} & 11.62 & 450.71 & 76.57\\
    \bottomrule
\end{tabular}
  }
\end{table*} 
We consider a HFEL system with 72 devices and 8 servers (clusters). Each cluster has 9 devices and 1 server. In the experiments, we employ three image classification datasets: FEMNIST \cite{caldas2019leaf}, CIFAR-10 \cite{krizhevsky2009learning}, and FMNIST\cite{xiao2017/online}. The FEMNIST dataset is the federated splitting version of the EMNIST dataset, which includes 3,550 writers. We randomly sample 72 writers to simulate the practical HFEL application. we divide each writer's local data into $90\%$ for training and $10\%$ for testing. For evaluation purposes, a common testing dataset is constructed by aggregating the testing data from all devices. Note that FEMNIST is a non-IID distributed dataset due to the variations in writing styles among different writers. We train a modified ResNet-20 for FEMNIST, comprising a total of 272,814 parameters.

The CIFAR-10 dataset is composed of 50,000 training images and 10,000 testing images. To simulate the practical data distribution scenarios, we adopt three data partition strategies: IID distribution, Dirichlet distribution, and cluster-pathological distribution. In the case of IID distribution, the 50,000 training images are evenly and randomly distributed among all the devices, ensuring each device receives an equal share of the dataset. For Dirichlet distribution, the partitioning of the 50,000 training images among devices follows the Dirichlet distribution \cite{hsu2019measuring} with a concentration parameter of 1. For cluster-pathological distribution, we first partition the 50,000 training images into 8 clusters based on the given number of labels per cluster (LC). Within each cluster, the data are distributed among all the devices in an IID fashion, similar to the IID distribution strategy. This method aims to capture the characteristics of clustered data in real-world scenarios. We train a ResNet-20 (269,722 parameters) on CIFAR-10. The original 10,000 testing images are used as the common testing dataset. 

The FMNIST dataset comprises 60,000 training images and 10,000 testing images. Similar to the CIFAR-10 dataset, we utilize three data partitioning strategies in the experiments. We train a LeNet-5 (431,080 parameters) on FMNIST. The original 10,000 testing images are used as the common testing dataset. To demonstrate the effectiveness of FedRT, we compare it with three baselines: \emph{Static-R}, \emph{Static-T}, and \emph{CE-FedAvg}. The details of baseline algorithms are as follows:
\begin{itemize}
    \item \emph{Static-R} assumes that each edge server evenly assigns its communication bandwidth among all edge devices within its cluster. To fulfill the energy requirements of the HFEL system, we solve the optimization problem as in FedRT to determine the feasible CPU frequency and the edge backhaul topology, except the allocated communication bandwidth remains constant.
    
    \item \emph{Static-T} adopts a static edge backhaul topology and only considers the resource allocation for edge devices. To satisfy the energy constraint, the communication bandwidth and CPU frequency are optimized following the same methodology as FedRT, except the edge backhaul topology is fixed. 
    
    \item \emph{Ce-FedAvg~\cite{zhang2022scalable}} assumes that all edge devices are homogeneous and the edge backhaul topology is given a prior. We adapt it to our problem setting by employing the fixed edge backhaul topology and static resource allocation. Specifically, the server communicates following the base edge backhaul topology. The communication bandwidth is evenly assigned among edge devices. To satisfy the energy requirement, similar to \emph{Static-R} and \emph{Static-T}, we resolve the optimization problem as in FedRT to obtain a feasible CPU frequency for each edge device.

    \item \emph{MLL-SGD}~\cite{castiglia2021multilevel} adaptively assigns different local training iterations to each device based on their resource capabilities. For instance, slower devices are assigned fewer training steps to save computation time, while faster devices receive more steps to maintain model performance. In adapting this approach to our problem setting, we set different numbers of local training iterations according to each edge device's computation resource. Similar to Ce-FedAvg, we solve the optimization as FedRT with fixed edge backhaul topology and homogeneous communication bandwidth to derive the feasible CPU frequency for edge devices.

\end{itemize}
    
    

\subsection{Datasets and Models} 
For all experiments, we use mini-batch SGD with 0.9 momentum to train the local model with a batch size of 32. The learning rate is 0.1 for CIAFR-10 and FEMNIST. For FEMNIST, the learning rate is 0.01. The number of local iterations is 10 for all datasets. The number of edge rounds is 2 for all experiments. For CIAFR-10, we run 300 global rounds under IID data distribution and 350 rounds under non-IID data distribution (Dirichlet distribution and cluster-pathological distribution). For FEMNIST, the total number of global rounds is 200. For FMNIST, the total number of global rounds is 100 under the IID data distribution and 150 under the non-IID distribution. We run each experiment with 3 random seeds and report the average. The number of server communication times in each global round is $\psi = 10$. 

We record the total completion time and testing accuracy for evaluation. We use thop\footnote{https://pypi.org/project/thop/} to estimate the computation workload in terms of the number of floating point operations (FLOPs). The number of FLOPs needed for each training sample per iteration is 123.9 MFLOPs for ResNet-20 on CIFAR-10, 94.2 MFLOPs for ResNet-20 on FEMNIST, and 3.9 MFLOPs for LeNet-5 on FMNIST, respectively. The device's communication power is set to 0.01 $\si{\watt}$. The maximum available CPU frequency $f_{n}^{\text{max}}$ is $3.0 \si{GHz}$ and the minimum CPU frequency $f_{n}^{\text{min}}$ is $2.0 \si{ GHz}$. To mimic the CPU capability of practical devices, we generate the efficient capacity coefficient for each device by multiplying a constant coefficient and a random variable. We adopt the value of efficient capacity coefficient from~\cite{yang2020energy} $\alpha_{n} = 2\times 10^{-28}$ as a constant coefficient, and the random variable is uniformly distributed within range $[0.01,0.1]$. The total available communication bandwidth $B = 1\si{MHz}$ for all servers. We assume the $\text{SNR}$ are uniformly distributed within the range $0-15$ \si{dB} for all devices and all time slots. The available energy supply is $\bar{\mathcal{E}}_{n}=1 \si{J}$ for all devices. By default, we assume the base graph topology is fully connected if no other topology is explicitly specified. To simulate the practical edge backhaul communication process, we assume $\mathbf{B}^t$ are randomly (uniformly) fluctuating within the range $0.1-10$ Mbps.

\subsection{Experimental Results}

\textbf{Performance Comparison with Baselines.} We first evaluate the convergence speed of FedRT in comparison to the baseline methods. Fig.~\ref{fig:baseline_cifar} shows the testing accuracy and training loss w.r.t training time (in seconds) on CIFAR-10 under both IID and Dirichlet data distribution. The results indicate that FedRT exhibits a faster convergence speed than baselines. Throughout the training process, FedRT consistently achieves higher testing accuracy and lower training loss under both IID and non-IID data distributions under the same training time budget. Moreover, FedRT reduces the total training latency on CIFAR-10 by $30.90\%$ and $31.14\%$ for IID and non-IID distributions, respectively. Fig.~\ref{fig:baseline_fmnist} depicts the comparison results on FMNIST under IID and non-IID data distribution. Fig.~\ref{fig:baseline_femnist} presents the comparison results on FEMNIST. Similarly, FedRT has a faster convergence speed than baselines. On the FMNIST dataset, the total training latency of FedRT is reduced by $36.43\%$ and $35.82\%$ under IID and non-IID data distribution, respectively. Moreover, the total training latency is reduced by $32.35\%$ compared to baselines on the FEMNIST dataset. The results of Fig.~\ref{fig:baseline_cifar}, Fig.~\ref{fig:baseline_fmnist}, and Fig.~\ref{fig:baseline_femnist} demonstrate that FedRT consistently outperforms all baseline methods across both IID and non-IID data distributions, thereby demonstrating the effectiveness of the proposed method.


\textbf{Energy Consumption Comparsion.} We then compare the energy consumption of FedRT against several baselines. Fig.~\ref{fig:ene_cifar} shows the curve of accumulated energy consumption for edge devices w.r.t the training round on CIFAR-10 under both IID and Dirichlet data distribution. The results indicate that FedRT consistently consumes less energy than the Static-R, CE-FedAvg, and MLL-SGD throughout the training process. As the number of training rounds increases, the discrepancy between the curves becomes more pronounced, demonstrating that our algorithm consistently reduces energy costs, and its advantage becomes increasingly evident. Furthermore, the energy consumption pattern of FedRT closely aligns with that of Static-T, which also implements resource allocation for devices. This observation underscores the effectiveness of FedRT in optimizing resources for energy conservation. Moreover, we can find that FedRT consumes more energy under non-IID distribution than IID, highlighting the critical need for resource optimization under data heterogeneity. Fig.~\ref{fig:ene_fmnist} and Fig.~\ref{fig:ene_femnist} show the accumulated energy consumption w.r.t the training round on FMNIST and FEMNIST datasets. We have similar conversations about energy consumption. In conclusion, FedRT achieves better energy usage compared with Static-R, CE-FedAvg, and MLL-SGD baselines. Although the energy consumption of FedRT is nearly identical to that of Static-T, FedRT achieves a faster convergence rate, showing an advantage in training performance.

\textbf{Latency, Energy, and Accuracy.} Table.~\ref{compare-table} presents a comprehensive comparison of the total training time, total energy consumption, and the highest testing accuracy across all datasets and methods. The results clearly demonstrate that FedRT consistently exhibits the shortest training time, demonstrating the effectiveness of our method in minimizing training latency. Notably, FedRT achieves the highest final testing accuracy on the CIFAR-10 dataset under the IID distribution, the FMNIST dataset under the IID distribution, and the FEMNIST dataset. Although FedRT's final testing accuracy is marginally lower than the best baselines for CIFAR-10 under the non-IID distribution and FMNIST under the non-IID distribution, the differences are minimal—only $0.14\%$ and $0.06\%$ below the highest accuracies, respectively. These findings validate the effectiveness of the consensus distance constraint in \textbf{P2}. Additionally, FedRT demonstrates superior performance in terms of energy efficiency, yielding the lowest energy consumption for the FMNIST dataset and nearly the lowest for the CIFAR-10 and FEMNIST datasets. In summary, FedRT achieves the best trade-off between the total training latency and testing accuracy while satisfying the energy constraint.

\textbf{Effect of Inter-cluster non-IID Data.} We evaluate FedRT in combating the inter-cluster non-IID data distribution. The number of labels within each cluster is $2,4,6,8$ both for CIFAR-10 and FMNSIT. With more labels within the cluster, the data heterogeneity tends to be IID distributed. In Fig.~\ref{fig:lc_noniid}, we show the convergence rate of FedRT and the baselines without topology optimization (Static-T and CE-FedAvg) w.r.t. to global rounds. We can find the testing accuracy decreased as data heterogeneity increased. Compared with Static-T/CE-FedAvg, FedRT-RT has a best convergence speed.

\textbf{Effect of Base Graph.} We study the impact of base graph topology on the performance of FedRT. We generate the random graph as the base graph through Erdős–Rényi method~\cite{erdds1959random}. The probability for edge creation is set to be $0.2, 0.4, 0.6, 0.8, 1.0$. With a higher probability for edge creation, the graph topology is more densely connected. The probability is 1.0, which means the generated graph is a fully connected graph. 
We illustrate the convergence rate in terms of global rounds for CIFAR-10 and FEMNIST. The data distribution for CIFAR-10 is IID. From Fig.~\ref{fig:topo}, FedRT achieves a fast convergence rate under the IID data distribution for CIAFR-10. For FEMNIST, a loss connection leads to a slightly slow convergence rate. However, in most cases, FedRT yields a similar curve. This result demonstrates that FedRT works well under sparse base graph topology.

\section{Conclusion}\label{sec:conclusion}
This paper explored resource-efficient federated learning for a two-tier networked system. We formulated an optimization problem to minimize training latency under the given energy budget. We proposed an efficient algorithm, FedRT, to resolve the formulated problem. Experiment results proved the effectiveness of FedRT in reducing training latency and retaining high accuracy compared to conventional methods.
\bibliographystyle{IEEEtran}
\bibliography{main.bib}

\end{document}